\pdfoutput=1

\documentclass[11pt]{article}

\usepackage[final]{acl}

\usepackage{times}
\usepackage{latexsym}

\usepackage[T1]{fontenc}

\usepackage[utf8]{inputenc}
\usepackage{amsmath}
\usepackage{microtype}

\usepackage{inconsolata}

\usepackage{graphicx}
\usepackage{booktabs} 
\usepackage{algorithm}
\usepackage{algpseudocode}

\usepackage{enumitem}
\usepackage{mathtools}
\mathtoolsset{showonlyrefs}
\usepackage{multirow}
\usepackage[most]{tcolorbox}
\usepackage{url}
\usepackage{hyperref}
\usepackage{scalerel,graphicx,xparse}
\usepackage{microtype}
\usepackage{hyperref}
\usepackage{url}
\usepackage{booktabs}
\usepackage{graphicx}
\usepackage{lineno}
\usepackage{multirow}
\usepackage{amsmath} 
\usepackage{algorithm}
\usepackage{algpseudocode}
\usepackage{wrapfig}
\usepackage{caption}
\usepackage{subcaption}
\usepackage[most]{tcolorbox}
\definecolor{darkblue}{rgb}{0, 0, 0.5}
\hypersetup{colorlinks=true, citecolor=darkblue, linkcolor=darkblue, urlcolor=darkblue}

\title{Towards Robust Evaluation of Visual Activity Recognition: Resolving Verb Ambiguity with Sense Clustering}

\author{
  Louie Hong Yao$^{1}$ \qquad Nicholas Jarvis$^{2}$ \qquad Tianyu Jiang$^{2}$ \\
  $^{1}$Independent Researcher \qquad $^{2}$University of Cincinnati \\
  \texttt{lhyao731@gmail.com, jarvisns@mail.uc.edu, tianyu.jiang@uc.edu}
}

\begin{document}
\maketitle
\begin{abstract}
Evaluating visual activity recognition systems is challenging due to inherent ambiguities in verb semantics and image interpretation. When describing actions in images, synonymous verbs can refer to the same event (e.g., \textit{brushing} vs. \textit{grooming}), while different perspectives can lead to equally valid but distinct verb choices (e.g., \textit{piloting} vs. \textit{operating}). Standard exact-match evaluation, which relies on a single gold answer, fails to capture these ambiguities, resulting in an incomplete assessment of model performance. To address this, we propose a vision-language clustering framework that constructs \textbf{verb sense clusters}, providing a more robust evaluation. Our analysis of the \mbox{imSitu} dataset shows that each image maps to around four sense clusters, with each cluster representing a distinct perspective of the image. We evaluate multiple activity recognition models and compare our cluster-based evaluation with standard evaluation methods. Additionally, our human alignment analysis suggests that the cluster-based evaluation better aligns with human judgments, offering a more nuanced assessment of model performance.  
\end{abstract}

\section{Introduction}

\begin{figure}[t]
    \centering
    \includegraphics[width=0.98\linewidth]{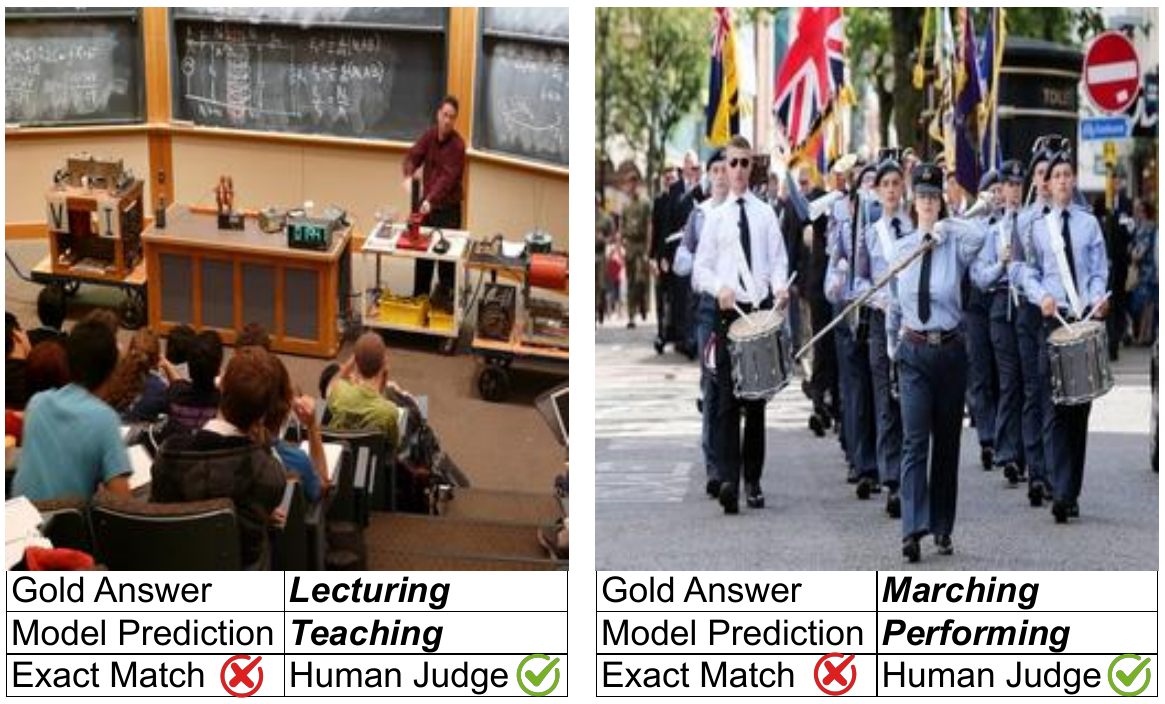}
    \caption{Accuracy score based on the exact match against gold labeling does not consider the ambiguity.}
    \label{fig:intro}
\end{figure}

Visual activity recognition systems aim to interpret events in images by predicting the primary activity, such as \textit{riding} and \textit{cooking}. However, evaluating these systems presents significant challenges due to multiple levels of ambiguities. One ambiguity comes from the complexity of word meanings and semantic relations. For instance, synonymous verbs can describe the same event: \textit{teaching} and \textit{lecturing} often refer to the same action. 
Another ambiguity lies within the images. When describing the event in one image from different perspectives, the choice of verbs can vary. For example, an image of a marching band can be described by both \textit{marching} and \textit{performing}. As shown in Figure~\ref{fig:intro}, a model should receive credit for predicting \textit{teaching}/\textit{performing} despite that the ground truth label is \textit{lecturing}/\textit{marching}.

Previous work on visual action recognition has been adopting discrete semantic classes to describe actions, both in static images~\citep{ronchi2015describing, chao2015hico} and videos~\citep{gu2018ava, wang2024hardvs}. \citet{yatskar2016situation} introduced the \textit{imSitu} dataset, which has become a widely adopted benchmark for visual activity recognition. The task is defined as, given an image, the model should first identify the appropriate verb describing the image, then fill in the other participants in the event. The imSitu dataset comprises 126,102 images annotated with 504 verbs, covering a broad spectrum of daily activities. Previous approaches to this task~\citep{cooray2020attention, li2022clip, jiang2023exploiting} have been following the standard evaluation based on exact match: a prediction is correct only when the output verb is the same as the gold answer. We argue that this evaluation metric does not reflect the true capability of a recognition system due to two types of ambiguity: (1) \textbf{Synonymy} - multiple synonymous verbs that describe the same activity, and (2) \textbf{Multi-perspectives} - multiple valid verbs that describe the activity from different perspectives.

To address the ambiguity in evaluation, we propose a novel vision-language clustering framework that constructs verb sense clusters. These clusters serve two purposes: (1) quantifying inherent ambiguity in exact-match evaluation and (2) enabling more accurate assessment of visual activity recognition systems. Using both internal and external metrics, we validate the effectiveness of our clustering approach. Based on our sense clusters, we reveal that on average, an image can typically be described using four sense clusters, and each cluster contains close to two  synonymous verbs. Through extensive experimentation comparing multiple supervised models and multimodal large language models, we demonstrate that our cluster-based evaluation methodology offers a more comprehensive and accurate assessment than traditional exact-match approaches. Our code is publicly available.\footnote{\url{https://github.com/ruyi101/multimodal-verb-senses}} In summary, our contributions are three-fold:
\vspace{-1.5mm}
\begin{enumerate}
    \item We propose a novel vision-language clustering framework to build sense groups based on images and activity verbs, and evaluate the clustering quality with both internal and external metrics. 
    \vspace{-1.5mm}
    \item We demonstrate that our cluster-based evaluation offers a more accurate and nuanced assessment than exact-match accuracy.
    \vspace{-1.5mm}
    \item We quantify two types of ambiguity in evaluation based on exact match using the \mbox{imSitu} dataset: synonymous verbs describing the same activity and multiple valid verbs capturing different perspectives.
\end{enumerate}
While our experiments focus on imSitu, the proposed clustering-based evaluation framework is designed to be readily adaptable to other visual activity datasets.

\section{Related Work}
Visual activity recognition has seen significant development across both image and video domains, with datasets of varying granularity and scope. Early examples of image-based datasets include Stanford-40~\citep{yao2011human}, which covers 40 common human activities, and its extension to 89 activities by \citet{le2013exploiting}. Subsequent datasets such as \mbox{COCO-a}~\citep{ronchi2015describing} and HICO~\citep{chao2015hico} focus on everyday actions with 140 and 117 classes respectively, while more comprehensive collections like imSitu~\citep{yatskar2016situation} feature 504 verbs drawn from FrameNet, and SVO-Probes~\citep{hendricks-nematzadeh-2021-probing} cover 421 verbs. Video datasets show similar diversity in their treatment of actions, from AVA’s 80 atomic visual actions~\citep{gu2018ava} to VidSitu’s~\citep{Sadhu_2021_CVPR} extensive coverage of 1,500 verbs from PropBank. More recently, HARDVS~\citep{wang2024hardvs} introduced 300 fine-grained activity categories captured using dynamic vision sensors. This broad spectrum of activity classes across datasets highlights both the complexity of action recognition and the potential utility of our cluster-based evaluation framework.

The imSitu dataset has served as a key benchmark for advancing visual activity recognition techniques. Early approaches employed CNN-based backbones for image encoding~\citep{yatskar2016situation,pratt2020grounded}. More recent work has leveraged CLIP-based models~\citep{li2022clip, roy2024clipsitu} and transformer-based models~\citep{cho2022collaborative, jiang2023exploiting}. While these approaches have progressively improved performance, their reliance on exact verb matching for evaluation may underestimate their true capabilities. Although CLIPScore~\citep{hessel2021clipscore} has gained popularity as a flexible evaluation metric for image captioning using CLIP embeddings, its generic similarity measures are not suited for the specific challenges of verb classification. Our work also relates to recent efforts in visual word sense disambiguation~\citep{raganato2023semeval}, which addresses word polysemy in visual contexts.

Semantic evaluation of vision-language models has traditionally relied on metrics like BLEU~\citep{papineni2002bleu}, METEOR~\citep{banerjee2005meteor}, and CIDEr~\citep{vedantam2015cider}, which emphasize n-gram overlap but often miss deeper semantic alignment. More recent metrics such as CLIPScore~\citep{hessel2021clipscore}, RefCLIPScore~\citep{hessel2021clipscore}, and UMIC~\citep{lee2021umic} leverage vision-language embeddings to better assess cross-modal relevance. However, these metrics remain limited in capturing fine-grained semantic distinctions, such as verb sense ambiguity and shifts in perspective.

\section{Methodology}
\begin{figure*}[t]
    \centering
\includegraphics[width=0.81\linewidth,height=0.23\textheight,keepaspectratio=false]{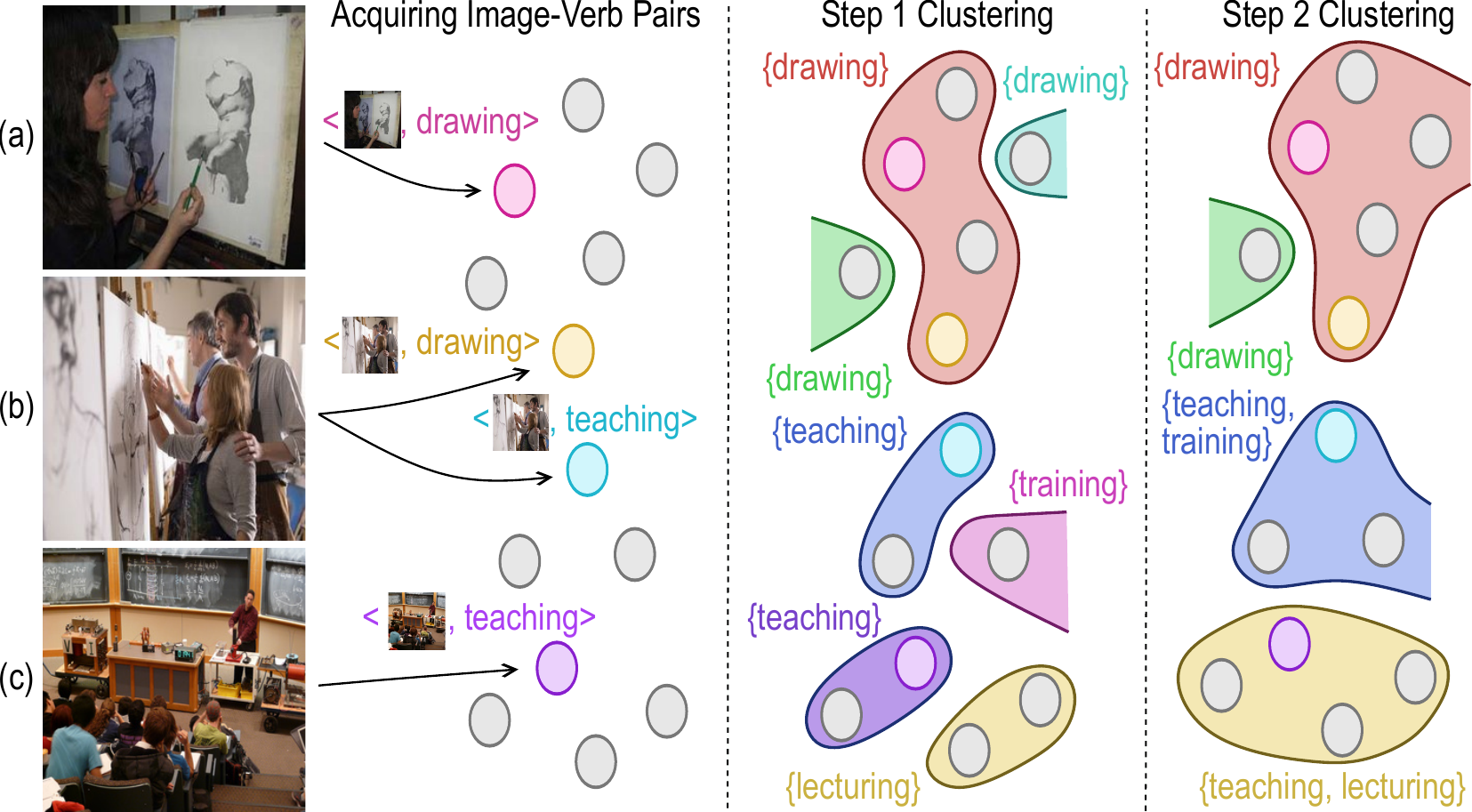}
    \caption{Overview of our two-step clustering framework.
Step 1 groups \texttt{<image,verb>} pairs with the same verb into fine-grained sense clusters. Step 2 merges these across verbs to capture shared semantic meanings.}
    \label{fig:clustering}
\end{figure*}

We propose a two-step clustering framework that clusters images into verb sense groups. We hypothesize that each cluster represents a fine-grained group of activities that are similar. For example, a cluster of different images depicting a teacher teaching in the classroom, with verbs \textit{lecturing}, \textit{instructing}, \textit{teaching}, \textit{educating} all describing this scenario. We use a multimodal large language model to generate image-verb pairs as nodes and cluster them  first by shared verbs, then into fine-grained sense groups. Figure~\ref{fig:clustering} shows our two-step clustering framework.

\subsection{Acquiring Image-Verb Pairs}
The original imSitu dataset collected the images by querying Google image search with verbs and using those verbs as gold labels for the returned images, which potentially introduces labeling bias. For example, searching for ``\textit{drawing}'' images might return both picture (a) and (b) in Figure~\ref{fig:clustering}. However, when considering the main activity in picture (b), ``\textit{drawing}'' and ``\textit{teaching}'' are equally valid answers.  To mitigate this bias, we query two multimodal large language models to generate all appropriate verbs to describe each image: GPT-4o mini~\citep{hurst2024gpt} and Llama-3.2-90B~\citep{dubey2024llama}. Prompting details can be found in Appendix~\ref{app:prompt}.

For each image, we collect a set of verbs through prompting and include the original gold-label verb if it is not already present. We then filter these verbs to retain only those that appear in the \mbox{imSitu} dataset. Next, we construct nodes, where each node represents an \texttt{<image,verb>} pair. To capture the joint semantic meaning of the image and verb, we use a multimodal Llama model to generate embeddings by processing both inputs simultaneously. Specifically, we extract the final hidden state after the text-image cross-attention layer and use it as the representation vector for each node.

\subsection{Two-Step Clustering}
The two-step clustering framework aims to group \texttt{<image,verb>} pairs into coherent clusters, so that each cluster represents one type of activities with similar images and verbs. This subsection introduces the procedure, with the corresponding pseudo-code provided in Algorithm~\ref{alg:clustering} in the appendix.

\vspace{3pt}
\noindent \textbf{Step 1: Same-Verb Clustering.} In the first step, the goal is to disambiguate the fine-grained senses of each individual verb. For each verb $v$ in the list of 504 target verbs $\mathcal{V}$, we first collect all images $\mathcal{I}_v$ associated with $v$ from the dataset $\mathcal{D}$. To capture both the visual and semantic properties of the \texttt{<image,verb>} pairs, the images and verbs are transformed into high-dimensional embeddings $\mathcal{E}_v$ using Llama-3.2-11B. Specifically, the verb $v$ is provided as input text, while the corresponding image is provided as the input image. The embeddings are extracted by leveraging the model's cross-attention mechanism, which integrates the semantic information from the verb and the visual information from the image. The final embedding for each \texttt{<image,verb>} pair is obtained from the last token of the model's final layer, which is a 4096-dimensional representation.

To cluster the embeddings $\mathcal{E}_v$, we apply two clustering algorithms, K-Means and Hierarchical Agglomerative Clustering (HAC), collectively referred to as $\mathcal{A}$. For K-Means, we normalize the embeddings to unit vectors so that Euclidean distance is monotonically related to cosine distance. 
When using HAC, we employ complete linkage and cosine distance to group the embeddings.  The optimal number of clusters $k$ for each verb is determined by exploring a candidate range of cluster numbers $\mathcal{K} = {2, 3, \dots, 16}$. For each $k \in \mathcal{K}$, the algorithm $\mathcal{A}$ generates a clustering result $\mathcal{C}_{k}$, which is evaluated using the Silhouette score. The $k$ that achieves the highest Silhouette score is selected as the optimal number of clusters for $v$, and the corresponding clustering result $\mathcal{C}_k$ is stored.

\vspace{3pt}
\noindent \textbf{Step 2: Cross-Verb Clustering.} The second step addresses cross-verb ambiguities by clustering the results from Step 1 into final clusters. The clustering algorithm $\mathcal{A}$ used in this step is the same as in Step 1. For example, if K-Means is used for clustering in Step 1, K-Means will also be used in Step 2. These final clusters represent shared meanings across verbs and provide a higher-level understanding of the actions depicted in the images.

Each cluster $c \in \mathcal{C}_{\text{step1}}$ is represented by its average embedding $\mathbf{e}_c$, which captures the semantic and visual properties of the cluster. The number of final clusters is determined by computing $k_r = \text{int}(504 \times r)$, where 504 is the total number of target verbs and $r$ is a ratio selected from the candidate list $\mathcal{R} = \{0.6, 0.7, \dots, 1.6\}$ with an increment of 0.1. The clustering algorithm $\mathcal{A}$ is then applied to the set of average embeddings $\{\mathbf{e}_c\}$ using $k_r$ clusters, producing the final clustering result $\mathcal{C}_{
\text{final}}$. 
The optimal hyperparameter $r$, and consequently $k_r$, is selected based on the Silhouette score. The clustering result $\mathcal{C}_{\text{final}}$ with the highest Silhouette score is chosen as the final output.

\section{Evaluation and Analysis}
We organize our analysis into four parts. We begin by evaluating the quality of our clustering framework and use the resulting clusters to analyze sources of ambiguity in the imSitu dataset. Next, we apply these clusters to evaluate a range of visual activity recognition systems. We further examine the robustness of our evaluation method and its alignment with human judgment. In the third part, we provide a qualitative analysis of the learned clusters. Finally, we assess the limitations of CLIPScore as an evaluation metric for verb-centric activity recognition.

\subsection{Quality of Clusterings and Ambiguity Analysis}
We begin by evaluating the quality of our clustering results, and then leverage them to analyze sources of ambiguity in the imSitu dataset.

\vspace{3pt}
\noindent
\textbf{Assessing the Quality of Clusters.} 
To assess the quality of our clustering, we use two internal metrics and one external diagnostic. As internal metrics, we report the Silhouette score~\citep{rousseeuw1987silhouettes} and the Calinski--Harabasz index~\citep{calinski1974dendrite}, which quantify cluster separation and cohesion. As an external reference, we compute a purity score with respect to WordNet synsets~\citep{miller1995wordnet}. For each cluster, we identify all synsets containing at least one verb in the cluster and compute purity using the maximum synset overlap (details in Appendix~\ref{app:clustering_metrics}). 

Importantly, while WordNet synsets encode \emph{linguistic senses} defined by lexical-semantic distinctions, our goal is to capture \emph{visually grounded senses} that reflect how actions are manifested and perceived in images. For instance, verbs such as \emph{running}, \emph{racing}, and \emph{chasing} correspond to distinct WordNet senses, yet they often share highly similar visual configurations involving motion, posture, and spatial relations. Accordingly, WordNet purity contributes to our assessment of clustering quality alongside internal metrics, serving as an auxiliary external diagnostic to detect over-merging of clearly unrelated verb meanings, rather than as a standalone measure of cluster quality.

\begin{table}[t]
    \centering
    \setlength{\tabcolsep}{8pt}
    \renewcommand{\arraystretch}{1.1}
    \resizebox{0.99\linewidth}{!}{
    \begin{tabular}{lccc}
        \toprule
        \textbf{Clustering Approach} & \textbf{Sil.} & \textbf{C-H} & \textbf{PuS} \\ 
        \midrule
        Llama-3.2-90B + K-Means     & \textbf{0.22} & 6.53 & \textbf{0.50} \\
        Llama-3.2-90B + HAC         & 0.18 & 5.33 & 0.40 \\
        GPT-4o mini + K-Means   & \textbf{0.22} & \textbf{8.05} & 0.45 \\ 
        GPT-4o mini + HAC       & 0.20 & 6.15 & 0.42 \\
        \bottomrule
    \end{tabular}
    }
    \caption{Clustering quality (Sil. = Silhouette, C-H = Calinski-Harabasz, PuS = Purity Score).}
    \label{tab:clustering_approaches}
\end{table}

In Table~\ref{tab:clustering_approaches}, we compare clustering results obtained by prompting GPT-4o mini and Llama-3.2-90B, using K-Means and hierarchical agglomerative clustering (HAC). K-Means consistently outperforms HAC across internal metrics, indicating stronger cluster cohesion and separation. While GPT-4o mini achieves higher Calinski--Harabasz scores under K-Means, Llama-3.2-90B yields higher WordNet purity. Although purity is not expected to be perfect, it provides a useful external reference when considered alongside internal metrics. Balancing internal cohesion with this external validation signal, we select the K-Means clustering derived from Llama-3.2-90B for evaluating visual activity recognition systems in Section~\ref{sec:evaluating_models}.

\begin{table}[t]
    \centering
    \setlength{\tabcolsep}{4pt}
    \renewcommand{\arraystretch}{1.1}
    \resizebox{0.98\linewidth}{!}{
    \begin{tabular}{lcc}
        \toprule
        \textbf{Metric} & \textbf{GPT-4o mini} & \textbf{Llama-3.2-90B} \\ 
        \midrule
        \# of Clusters         & 554   & 655   \\
        Verbs per Cluster      & 1.97  & 1.59  \\
        Clusters per Image     & 4.44  & 3.61  \\ 
        Multi-Image Rate       & 0.95  & 0.71  \\
        Clusters per Verb      & 2.17  & 2.07  \\
        Multi-Verb Rate        & 0.54  & 0.54  \\ 
        \bottomrule
    \end{tabular}
    }
    \caption{Clustering metrics illustrating ambiguity.}
    \label{tab:ambiguity_metrics}
\end{table}

\vspace{3pt}
\noindent
\textbf{Ambiguity Analysis.} 
Our clusters can also be used to analyze ambiguities within the imSitu dataset. We use the clusters derived from both GPT-4o mini and Llama-3.2-90B with K-Means to quantify ambiguity in verb usage. Table~\ref{tab:ambiguity_metrics} presents metrics that capture two types of ambiguity:

\smallskip
\begin{itemize}[topsep=0pt, itemsep=2pt]
  \item \textit{Synonymy.} It is captured when different verbs are grouped into the same cluster. Row 2 (Verbs per Cluster) shows that each cluster contains 1.59 (Llama) and 1.97 (GPT) verbs on average, indicating that synonyms are common within the 504 target set.
  \item  \textit{Multi-Perspectives.}  Verbs interpreted from different perspectives in a single image are identified when an image appears in multiple clusters. Both clustering results show that over 70\% of images belong to more than one cluster (Row 4, Multi-Image Rate), with each image appearing in 3.61 (Llama) and 4.44 (GPT) clusters on average (Row 3, Clusters per Image), i.e., about 4 perspectives per image. 

\end{itemize}

\smallskip

Despite differences in the number of final clusters produced by each model, the consistency observed across other metrics underscores the robustness of our evaluation framework.

\textit{Polysemy}, where a verb has multiple senses, though does not affect the model evaluations, can also be analyzed through clustering results. It is identified when a verb appears in multiple clusters. As shown in Table~\ref{tab:ambiguity_metrics}, over $50\%$ of the verbs span multiple clusters (Row 6, Multi-Verb Rate), indicating they represent different senses. On average, each verb is associated with 2 clusters (Row 5, Clusters per Verb).

\subsection{Evaluation of Visual Activity Recognition Systems}
\label{sec:evaluating_models}
\textbf{Model Evaluation.} The primary goal of our clustering approach is to enable more accurate evaluation of visual activity recognition systems. We validate this framework by evaluating multiple models and comparing their performance under three criteria: exact-match accuracy, WordNet-synset-based accuracy, and our cluster-based accuracy. In the WordNet-based evaluation, a prediction is considered correct if the predicted verb shares at least one synset with the gold target verb. For our cluster-based evaluation, we examine all clusters containing the target image; a prediction is deemed correct if the predicted verb appears in any of those clusters. Otherwise, it is marked incorrect. Our evaluation focuses on two types of systems:

\smallskip
\begin{itemize}[topsep=0pt, itemsep=2pt]
  \item \textit{Supervised Models.} We built a straightforward supervised architecture consisting of an image encoder backbone followed by a linear classification layer. We experimented with three image encoders: ResNet-50~\citep{he2016deep}, CLIP~\citep{radford2021learning}, and Llama-3.2-11B~\citep{dubey2024llama}. Although achieving state-of-the-art performance was not our primary objective, our Llama-11B-based model achieved strong results with a Top-1 accuracy of $56\%$ on the imSitu test set, comparable to the current SOTA (details in Appendix~\ref{app:supervised_models}).

  \item \textit{Multimodal LLMs.} To demonstrate the zero-shot capability of multimodal LLMs, we tested three models: GPT-4o mini, Llama-3.2-11B and Llama-3.2-90B. We designed two prompt settings: one providing 504 verbs to choose from (\textbf{closed}) and one without (\textbf{open}).
\end{itemize}
\smallskip

Table~\ref{tab:prediction_models_single_column} presents Top-1 and Top-5 accuracy under three evaluation criteria: exact match with the gold verb, WordNet-Synset (WN-Syn) match, and our cluster-based evaluation. The top section shows that models using Llama as the backbone achieve the best performance across all metrics. While all models benefit from cluster-based evaluation, CLIP and Llama exhibit larger gains compared to ResNet, suggesting that multimodal training leads to stronger semantic understanding.

\begin{table*}[t] 
    \centering
    \small
    \setlength{\tabcolsep}{12pt}
    \begin{tabular}{llcccccc}
        \toprule
         & & 
        \multicolumn{3}{c}{\textbf{Top-1 Accuracy}} & 
        \multicolumn{3}{c}{\textbf{Top-5 Accuracy}} \\ 
        \cmidrule(lr){3-5} \cmidrule(lr){6-8}
         &  \textbf{Model} & \textbf{Gold} & \textbf{WN-Syn} & \textbf{Cluster} & \textbf{Gold} & \textbf{WN-Syn} & \textbf{Cluster} \\ 
        \midrule
        &ResNet-50         & 30.5 & 32.5 & 43.2 & 55.9 & 58.2 & 70.7 \\ 
        &CLIP           & 46.2 & 48.4 & 61.5 & 75.5 & 76.9 & 86.7 \\ 
        &Llama-3.2-11B      & 56.0 & 58.4 & 72.1 & 83.1 & 84.3 & 91.8 \\ 
        \midrule
        \multirow{3}{*}{\it closed}
        &GPT-4o mini            & 17.3 & 20.5 & 39.2 & 55.7 & 60.2 & 85.3 \\ 
        &Llama-3.2-11B      & 25.3 & 28.6 & 52.4 & 37.7 & 41.2 & 69.8 \\ 
        &Llama-3.2-90B$^*$  & 24.9 & 27.4 & -    & 37.0 & 40.2 & -    \\ 
        \midrule
        \multirow{3}{*}{\it open}
        &GPT-4o mini            & 20.0 & 22.2 & 44.4 & 45.6 & 49.3 & 78.2 \\
        &Llama-3.2-11B      & 7.4 & 8.7 & 23.5 & 20.3 & 22.3 & 41.7 \\ 
        &Llama-3.2-90B      & 17.2 & 29.8 & 48.9 & 41.8 & 45.4 & 80.6 \\ 
        \bottomrule
    \end{tabular}
    \caption{Comparison of model performance using different evaluation methods. Top-1 and Top-5 accuracies are reported based on three criteria: exact match with the gold answer (Gold), WordNet synset-based evaluation (WN-Syn), and clustering-based evaluation (Cluster). The \textit{closed} setting provides 504 verbs to the LLMs in the prompt, while the \textit{open} setting does not. $^*$The clusters are derived from Llama-3.2-90B (closed) and therefore not used when evaluating that same configuration.}

    \label{tab:prediction_models_single_column}
\end{table*}

Zero-shot LLMs perform worse than supervised models under exact-match evaluation. However, their scores improve significantly with our cluster-based metric. This suggests their predictions, while not always matching the gold verb, are often semantically appropriate. 
We also find WordNet-based evaluation yields about a 2\% gain in Top-1 accuracy compared to exact match (56.0\%$\rightarrow$58.4\%). This gain reflects improvements from synonymy and aligns with the increase observed in our model accuracy improvement breakdown.

\begin{table}[t]
    \centering
    \resizebox{0.99\linewidth}{!}{
    \setlength{\tabcolsep}{5pt}
    \begin{tabular}{llcccc}
        \toprule
         & \textbf{Model} & \textbf{Gold} & \textbf{Cluster} & \textbf{Syn} & \textbf{Multi-P} \\ 
        \midrule
         & ResNet-50         & 30.5 & 43.2 & +3.9  & +8.8  \\ 
         & CLIP           & 46.2 & 61.5 & +4.9  & +10.4 \\ 
         & Llama-11B          & 56.0 & 72.1 & +5.4  & +10.7 \\ 
        \midrule
        \multirow{2}{*}{\textit{closed}} 
         & GPT-4o mini            & 17.3 & 39.2 & +5.0  & +17.0 \\ 
         & Llama-11B      & 25.3 & 52.4 & +6.0  & +21.2 \\ 
        \midrule
        \multirow{3}{*}{\textit{open}} 
         & GPT-4o mini            & 20.0 & 44.4 & +5.1  & +19.4 \\ 
         & Llama-11B      & 7.4  & 23.5 & +2.3  & +13.7 \\ 
         & Llama-90B      & 17.2 & 48.9 & +4.6  & +27.1 \\ 
        \bottomrule
    \end{tabular}
    }
    \caption{Top-1 accuracy improvements using cluster-based targets. \textit{Syn}: gain from synonyms. \textit{Multi-P}: gain from multi-perspective understanding.}
    \label{tab:improvement_breakdown}
\end{table}

\vspace{3pt}
\noindent
\textbf{Improvement Breakdown.}
In Table~\ref{tab:improvement_breakdown}, we analyze the breakdown of the improvements in Top-1 accuracy in all models, examining how much of the improvement is attributed to synonyms and how much is due to different perspectives of the image. 
To distinguish between these cases, we identify whether the Top-1 label, which appears in the target verb clusters but is not the gold target, is in the same cluster as the gold target. If it appears in the same cluster, the prediction is classified as a synonym of the gold target. Otherwise, it is considered a description of the image from a different perspective.


Table~\ref{tab:improvement_breakdown} shows that models more often predict verbs from different perspectives (\textit{Multi-P} column). For supervised classification models, addressing perspective-related challenges contributes improvements of 8\% to 10\%, accounting for a significant portion of the total improvement (12\% to 16\%). In large language models (LLMs), perspective-related adjustments yield even greater gains, with improvements as high as 27\%, compared to total improvements ranging from 22\% to 32\%. While models also face difficulties in predicting synonyms, the impact is less pronounced than the challenges posed by differing perspectives.

\begin{table*}[t]
    \centering
    \resizebox{\linewidth}{!}{
    \begin{tabular}{llllllcllllc}
        \toprule
         & 
         & \multicolumn{5}{c}{\textbf{Top-1 Accuracy}} 
         & \multicolumn{5}{c}{\textbf{Top-5 Accuracy}} \\
         \cmidrule(lr){3-7} \cmidrule(lr){8-12}
         & \textbf{Model} 
         & \textbf{Gold} 
         & \textbf{4o} 
         & \textbf{4o-mini} 
         & \textbf{Cluster} 
         & \textbf{Human} 
         & \textbf{Gold} 
         & \textbf{4o} 
         & \textbf{4o-mini} 
         & \textbf{Cluster} 
         & \textbf{Human} \\
        \midrule
        &ResNet-50        & $34_{(-15)}$ & $57_{(+8)}$ & $63_{(+14)}$  & $\textbf{47}_{(-2)}$ & 49 & $58_{(-21)}$ & $86_{(+7)}$ & $86_{(+7)}$   & $\textbf{77}_{(-2)}$ & 79 \\ 
        &CLIP          & $50_{(-20)}$ & $77_{(+7)}$ & $84_{(+14)}$   & $\textbf{64}_{(-6)}$ & 70 & $79_{(-15)}$ & $\textbf{94}_{(0)}$ & $98_{(+4)}$   & $89_{(-5)}$ & 94 \\ 
        &Llama-3.2-11B         & $62_{(-20)}$ & $90_{(+8)}$ & $88_{(+6)}$  & $\textbf{79}_{(-3)}$ & 82 & $81_{(-15)}$ & $100_{(+4)}$ & $99_{(+3)}$  & $\textbf{94}_{(-2)}$ & 96 \\ 
        \midrule
        \multirow{2}{*}{\rotatebox[origin=c]{0}{closed}}
        &GPT-4o mini           & $16_{(-40)}$ & $75_{(+19)}$ & $78_{(+22)}$  & $\textbf{39}_{(-17)}$ & 56 & $52_{(-35)}$ & $98_{(+11)}$ & $98_{(+11)}$  & $\textbf{85}_{(-2)}$ & 87 \\ 
        &Llama-3.2-11B     & $16_{(-31)}$ & $\textbf{52}_{(+5)}$ & $\textbf{52}_{(+5)}$   & $39_{(-8)}$  & 47 & $20_{(-38)}$ & $83_{(+25)}$ & $82_{(+24)}$  & $\textbf{61}_{(+3)}$ & 58 \\ 
        \midrule
        \multirow{3}{*}{\rotatebox[origin=c]{0}{open}}
        &GPT-4o mini           & $18_{(-43)}$ & $80_{(+19)}$ & $77_{(+16)}$  & $\textbf{48}_{(-13)}$ & 61 & $44_{(-38)}$ & $95_{(+13)}$ & $97_{(+15)}$  & $\textbf{76}_{(-6)}$ & 82 \\
        &Llama-3.2-11B     & $6_{(-20)}$  & $38_{(+12)}$ & $36_{(+10)}$   & $\textbf{23}_{(-3)}$  & 26 & $16_{(-24)}$ & $74_{(+34)}$ & $84_{(+44)}$   & $\textbf{41}_{(+1)}$ & 40 \\ 
        &Llama-3.2-90B     & $12_{(-43)}$ & $75_{(+20)}$ & $70_{(+15)}$  & $\textbf{55}_{(0)}$   & 55 & $34_{(-47)}$ & $95_{(+14)}$ & $94_{(+13)}$  & $\textbf{80}_{(-1)}$ & 81 \\ 
        \bottomrule
    \end{tabular}
    }
    \caption{Comparison of Top-1 and Top-5 accuracy on 100 randomly sampled test images. Accuracy is measured using gold labels (exact match), GPT-4o and GPT-4o mini as judges, verb clusters, and human annotations. Subscripts show differences from human labels, and \textbf{bold} marks the closest value.}

    \label{tab:human_labeling}
\end{table*}

\begin{table}[t]
\centering
\small
\resizebox{0.99\linewidth}{!}{%
\begin{tabular}{lcccc}
\toprule
\textbf{Metric} 
& \multicolumn{2}{c}{\textbf{Kendall's $\tau$}} 
& \multicolumn{2}{c}{\textbf{Krippendorff's $\alpha$}} \\
\cmidrule(lr){2-3} \cmidrule(lr){4-5}
& \textbf{Top-1} & \textbf{Top-5}
& \textbf{Top-1} & \textbf{Top-5} \\
\midrule
Gold (exact match)  & 69.1 & 78.6 & 39.5 & 37.9 \\
CLIPScore-based     & 57.1 & 78.6 & 33.4 & 47.6 \\
Cluster (ours)      & 76.4 & 85.7 & 56.7 & \textbf{60.7} \\
\midrule
\multicolumn{5}{l}{\textit{GPT-4o as judge}} \\
r1 & 90.9 & 76.4 & \textbf{63.2} & 44.6 \\
r2 & 76.4 & 92.9 & 62.4 & 45.8 \\
r3 & \textbf{98.2} & 88.9 & 47.2 & 30.5 \\
r4 & \textbf{98.2} & \textbf{92.9} & 58.4 & 54.0 \\
\midrule
\multicolumn{5}{l}{\textit{GPT-4o mini as judge}} \\
r1 & 92.9 & 90.9 & 56.7 & 35.0 \\
r2 & 90.9 & 90.9 & 58.7 & 33.6 \\
r3 & 85.7 & 81.5 & 44.1 & 20.8 \\
r4 & 92.9 & 61.8 & 56.4 & 28.4 \\
\bottomrule
\end{tabular}
}
\caption{Agreement between evaluation metrics and human judgments on the 100-image subset.
We report Kendall’s $\tau$ and Krippendorff’s $\alpha$, with all values scaled by 100 for readability. For GPT-based judges, rounds r1–r2 use the same prompt as in Table~\ref{tab:human_labeling}, while r3–r4 use varied prompts to assess prompt sensitivity.}
\label{tab:agreement-human}
\end{table}


\vspace{3pt}
\noindent
\textbf{Human Judgment.}
To assess how well different evaluation strategies reflect human judgment, we analyze alignment from two complementary perspectives: \emph{absolute accuracy alignment} and \emph{relative agreement in ranking}. Implementation details for the evaluation setup are provided in Appendix~\ref{app:llm as judge}.

For absolute alignment, we randomly sampled 100 images and manually annotated the correctness of Top-1 and Top-5 predictions produced by all supervised models and LLMs. Table~\ref{tab:human_labeling} reports human-labeled accuracies alongside accuracies computed using gold labels, verb clusters, and VLM-as-a-judge (GPT-4o and GPT-4o mini). Exact-match evaluation against the gold verb systematically underestimates performance, reflecting verb semantic ambiguity, while VLM-as-a-judge consistently overestimates accuracy. In contrast, our cluster-based evaluation yields accuracy estimates that are closest to human judgments across models and settings.


Using the same 100-image subset, we further analyze relative alignment between evaluation metrics and human judgments by measuring Kendall’s $\tau$ (ranking agreement) and Krippendorff’s $\alpha$ (label agreement). Results are shown in Table~\ref{tab:agreement-human}. We compare our cluster-based evaluation against three alternatives, namely gold-label matching, VLM-as-a-judge, and a CLIPScore-based~\citep{hessel2021clipscore} method. For the CLIPScore criterion, we compute the CLIPScore between the predicted verb and the image, and deem the prediction correct if this score is greater than or equal to the CLIPScore between the gold verb and the same image. For VLM-as-a-judge, we report four independent rounds, each a separate run of the judge on the same images. Rounds~1-2 use the same prompt as in Table~\ref{tab:human_labeling} to assess sampling stochasticity. Rounds~3-4 use alternative prompts to evaluate robustness to prompt variation.

Our cluster-based evaluation achieves substantially higher agreement with human judgments than both exact match and the CLIPScore-based method across Top-1 and Top-5 evaluations. In contrast, VLM-as-a-judge exhibits notable instability, with agreement varying across repeated runs using the same prompt and fluctuating further when prompts are changed. Although GPT-4o and GPT-4o mini often preserve relative model ordering and thus attain high Kendall’s $\tau$ despite overestimating absolute performance, Krippendorff’s $\alpha$ offers a clearer view of absolute alignment by showing that our cluster-based evaluation aligns more reliably with human labels and exhibits lower variability.

\begin{figure*}[t]
   \centering
    \includegraphics[width=0.95\linewidth]{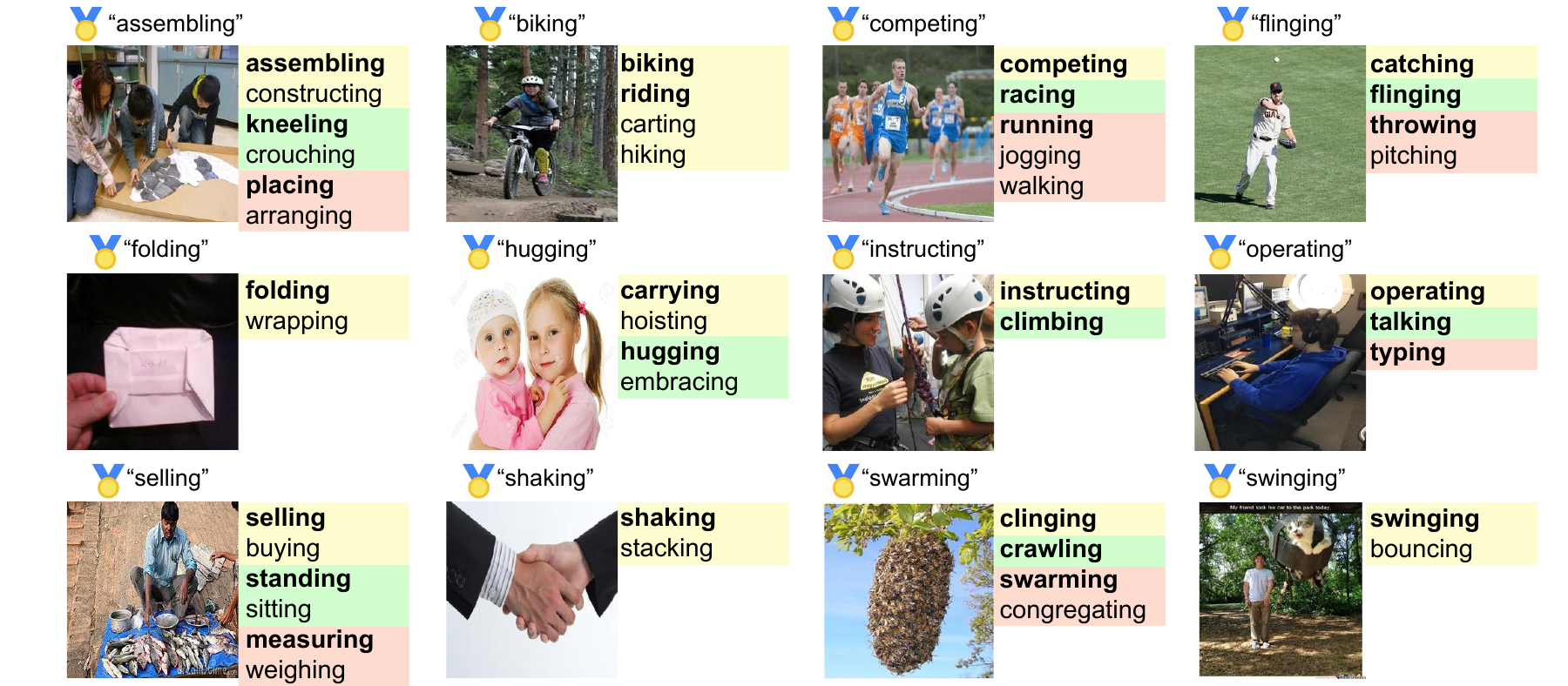}
    \caption{Qualitative examples of our clustering output. Each mini-panel shows: the imSitu gold verb (medal icon), Llama-3.2-90B predictions in \textbf{bold}, and additional verbs grouped in the same sense cluster. Different background colors indicate distinct clusters associated with the image.}
    \label{fig:error analysis Llama}
\end{figure*}

\begin{figure}[ht!]
    \centering
    \includegraphics[width=0.98\linewidth]{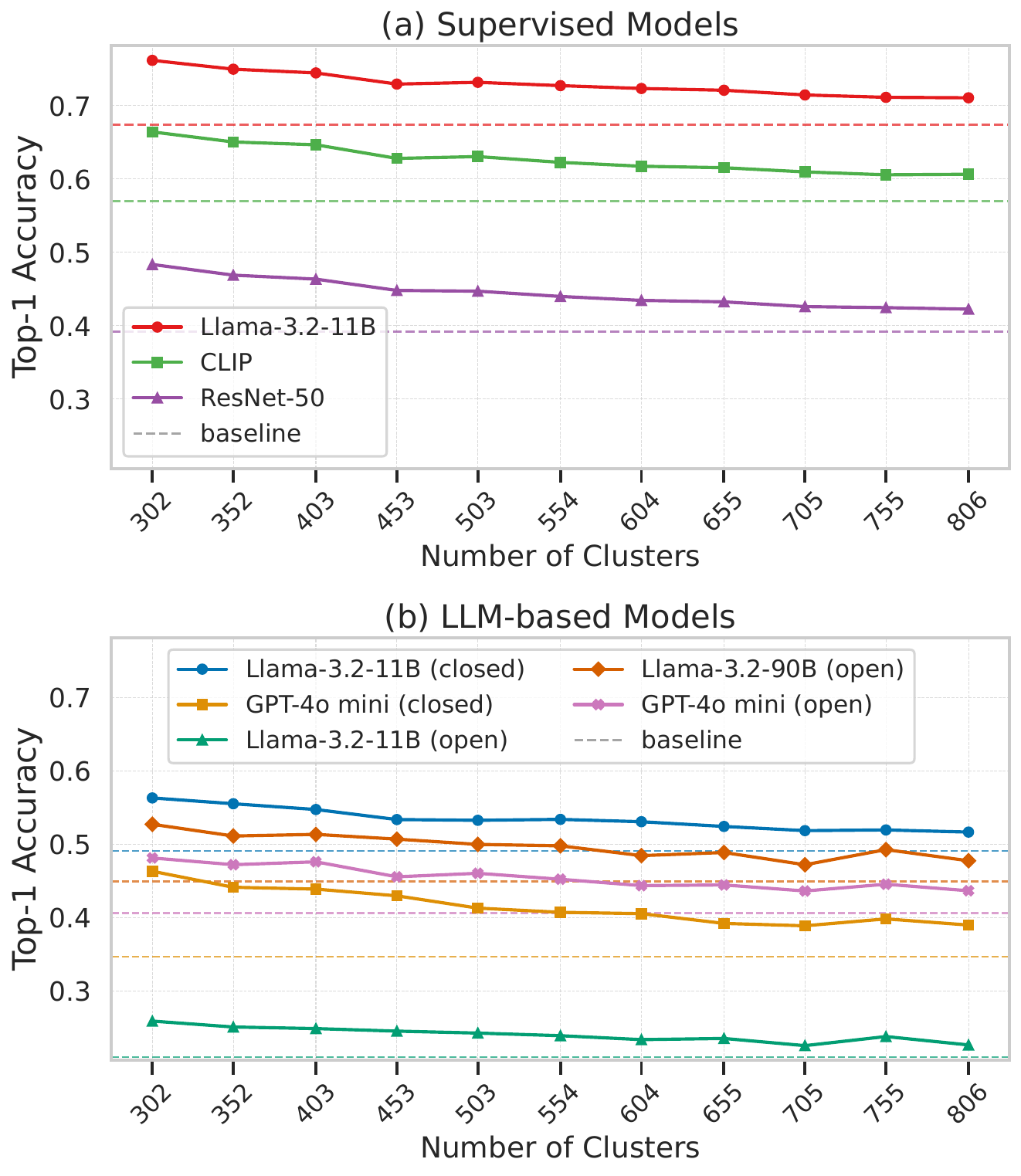}
    \caption{Top-1 accuracy of different models based on our clustering results. The dashed baselines represent the accuracy of the models when we use only the Llama-3.2-90B (closed) outputs as target verbs (no clustering), i.e., model's prediction verb is considered correct when it matches any of the targets.}
    \label{fig:robustness}
\end{figure}

\vspace{3pt}
\noindent
\textbf{Robustness to Cluster Count.}
We evaluate the robustness of our cluster-based framework by systematically varying cluster counts while holding other parameters constant. By re-assessing visual activity recognition models under these different cluster configurations, we examine how evaluation outcomes depend on cluster granularity. As shown in Figure~\ref{fig:robustness}, models' Top-1 accuracy demonstrates stable evaluation patterns across cluster counts. We maintain methodological consistency with our primary evaluation (Section~\ref{sec:evaluating_models}), using K-Means clustering of Llama-3.2-90B's closed-prompt verb predictions as our basis.

As expected, our analysis reveals a consistent inverse correlation between cluster granularity and model accuracy. This expected pattern emerges because finer clusters impose stricter semantic criteria, progressively limiting the number of valid target verbs per image. The accuracy decline stabilizes at approximately 655 clusters, which serves as the configuration for our main evaluations. 
To properly benchmark these results, we implement a baseline (visually denoted by a dashed horizontal line) using unprocessed Llama-3.2-90B outputs (closed) as direct reference data - the same outputs used for cluster generation. We count predictions as correct when they match any of the reference verbs. The performance gap between the baseline and our cluster-based evaluation demonstrates our method's crucial advantage: its inherent capacity to recognize linguistically valid variations, including synonymous expressions and semantically related verb forms, while maintaining rigorous evaluation standards.

\vspace{3pt}
\noindent
\textbf{Summary.}
We evaluate both supervised models and LLMs using our cluster-based framework, showing improved accuracy driven by both synonym resolution and multi-perspective understanding. The evaluation aligns more closely with human judgment, and robustness analysis confirms consistent accuracy trends across different clustering settings.

\subsection{Example of our Clustering Results}
To qualitatively demonstrate the results of our clustering, we present Figure~\ref{fig:error analysis Llama}. The 12 examples shown are randomly selected from the test set, each with its gold label (illustrated by the medal emoji), inference predictions (in \textbf{bold}), and other verbs within the same cluster. Different clusters associated with the same image are distinguished by background colors.

\begin{table*}[t]
    \centering
    {\renewcommand{\arraystretch}{1.1}
    \resizebox{0.95\linewidth}{!}{
    \begin{tabular}{cccccccc}
        \toprule
        \multirow{2}{*}{} &
        \multirow{2}{*}{\textbf{Statistics}} &
        \multirow{2}{*}{\textbf{Img-Verb$_{gold}$}} &
        \multirow{2}{*}{\textbf{Img-Cap$_{gold}$}} &
        \multicolumn{2}{c}{\textbf{GPT Adversarial}} &
        \multicolumn{2}{c}{\textbf{Random Choice}} \\
        \cmidrule(lr){5-6} \cmidrule(lr){7-8}
        &  &  &  &
        \textbf{Img-Cap$_{adv}$} & \textbf{Cap$_{gold}$-Cap$_{adv}$} &
        \textbf{Img-Cap$_{rand}$} & \textbf{Cap$_{gold}$-Cap$_{rand}$} \\
        \midrule
        \multirow{2}{*}{Dev}
            & Mean & 23.92 & 27.41 & 26.03 & 92.59 & 26.20 & 94.03 \\
            & Std  & 2.99  & 3.53  & 3.76  & 3.59  & 3.81  & 4.38 \\
        \midrule
        \multirow{2}{*}{Test}
            & Mean & 23.93 & 27.41 & 26.04 & 92.57 & 26.18 & 93.95 \\
            & Std  & 2.98  & 3.50  & 3.73  & 3.92  & 3.79  & 4.43 \\
        \bottomrule
    \end{tabular}
    }}
    \caption{CLIPScores (scaled by 100) for image-caption alignment using gold captions (Img-Cap$_{gold}$), GPT adversarial captions (Img-Cap$_{adv}$), and random-verb captions (Img-Cap$_{rand}$). We also report caption-caption CLIPScores (Cap$_{gold}$-Cap$_{adv}$, Cap$_{gold}$-Cap$_{rand}$) to quantify textual similarity under these perturbations, and include Img-Verb$_{gold}$, which measures the CLIPScore between each image and its gold verb as a baseline reference.}
    \label{tab:clipscore caption}
\end{table*}

Using the first image as an example, Llama-3.2-90B predicted \textit{assembling}, \textit{kneeling}, and \textit{placing}. After our clustering steps, \textit{assembling} was grouped with \textit{constructing}, \textit{kneeling} with \textit{crouching}, and \textit{placing} with \textit{arranging}. The fact that this image is associated with three distinct clusters supports the interpretation that multiple activities are occurring simultaneously: \textit{assembling} reflects the overall task, \textit{kneeling} describes the physical posture, and \textit{placing} captures the fine-grained action of positioning components.
While the three inference verbs in this case correspond to different activities, there are examples where multiple predictions express the same underlying action. For instance, in the \textit{biking} image shown in the second column, Llama predicted both \textit{biking} and \textit{riding}, which were correctly grouped into a single cluster by our method.

While our method performs well in most cases, minor errors could still arise in the pipeline, both from the Llama model’s inference and from our clustering. For example, in the \textit{biking} image, \textit{carting} was incorrectly included in the same cluster, even though it does not accurately describe the scene. In the \textit{selling} image at the end of the first column, the Llama model labeled the activity as \textit{standing}, while the person is clearly \textit{sitting}. Interestingly, our clustering partially corrects this by including \textit{sitting} in the target cluster. A more detailed error analysis is provided in Appendix~\ref{app:error analysis}.

\subsection{Analysis of CLIPScore Performance}
While discrete gold labels often miss the nuances of visual activity recognition, an open question is whether score-based image-text alignment metrics can provide a better alternative. We study this question using CLIPScore~\citep{hessel2021clipscore}, which measures semantic similarity via CLIP embeddings.

A natural instantiation is verb-only alignment, where we compare an image directly to a \emph{single verb} by computing the CLIPScore between the image and the predicted verb. To assess this setting, we compute CLIPScores between each image and all 504 imSitu verbs, rank verbs by score, and evaluate Top-1 and Top-5 accuracy against the gold labels. CLIPScore reaches a Top-1 accuracy of $\sim$25\%, which is plausible given the multi-perspective nature of images, but Top-5 remains below 50\%, suggesting difficulty in reliably retrieving even human-validated labels.
We conjecture that this limitation stems in part from CLIP being trained primarily on caption-like text rather than isolated action words, so the verb signal is weak and easily confounded.

Motivated by this, we next test whether CLIPScore behaves more reliably in a caption setting.
Our experiment on imSitu (Table~\ref{tab:clipscore caption}) examines how verb changes within otherwise identical captions affect CLIPScore. Using the labeled verb and its semantic roles, we generate correct captions for each image (see Appendix~\ref{app:clipscore} for details). From each caption, we construct two modified variants by replacing the verb with an implausible alternative using GPT-4o (e.g., \textit{The man kneads clay at the garage.} $\rightarrow$ \textit{The man eats clay at the garage.}) or by substituting it with a random verb from the imSitu verb list. While correct captions yield slightly higher CLIPScores, the differences are minimal even for semantically incorrect substitutions. The corresponding caption-caption CLIPScores remain high ($>92\%$), and Img-Cap$_{gold}$ is only slightly higher than Img-Verb$_{gold}$, suggesting that adding caption context only weakly strengthens the verb signal. Overall, this points to a limitation of CLIPScore for action-focused evaluation, because contextual words dominate the similarity signal and changes to the verb have little effect, a tendency also reported in textual-only settings~\citep{garcia2021exploring}.


\section{Conclusion}
In this work, we address the challenge of evaluating visual activity recognition systems by developing a two-step clustering framework that accounts for inherent ambiguities in verb semantics and image interpretation. Our analysis reveals that images in the imSitu dataset frequently involve both synonymous verbs describing the same activity and multiple valid interpretations from different perspectives. Through evaluating multiple recognition systems, including both supervised models and zero-shot multimodal LLMs, we show that our cluster-based evaluation yields higher accuracy scores compared to exact-match evaluation, which better reflects these models' capabilities in understanding visual activities, as validated by our manual analysis.

\section*{Limitations}

While our results demonstrate the effectiveness of cluster-based evaluation for visual activity recognition, several limitations remain. First, our experiments are conducted solely on the imSitu dataset. Although imSitu provides a well-established benchmark with rich verb annotations, evaluating the framework on additional datasets would further strengthen its generality. Importantly, our approach is not tied to imSitu-specific annotations. To adapt the framework to other datasets, such as VidSitu or datasets with verb-phrase labels, one can (i) collect model-generated verb or verb-phrase predictions for each visual instance, (ii) embed image-text pairs using a joint vision-language model, (iii) cluster the resulting representations using the same data-driven criteria, and (iv) evaluate model predictions against the resulting clusters. For verb-phrase labels, phrases can be embedded directly or decomposed into head verbs with modifiers, enabling the same clustering procedure to be applied. While this process requires dataset-specific preprocessing and validation, it provides a principled starting point for constructing visually grounded evaluation criteria beyond exact-match labels.

Second, our clustering relies on verb outputs generated by large multimodal models (e.g., Llama-3.2-90B and GPT-4o mini), rather than exhaustive human annotation. Although we manually inspected random samples to validate output quality, more comprehensive human evaluations or precision--recall analyses would further strengthen this component. In addition, our current framework treats all verbs within a cluster as equally valid, without modeling finer-grained contextual preferences. Incorporating weighted or context-dependent relationships between verbs is an important direction for future work.

Finally, our method introduces a nontrivial one-time computational cost for candidate generation, embedding, and clustering, and can be more expensive than a minimal VLM-as-a-judge baseline when evaluating a single system once. However, this cost is largely offline and amortized, since the procedure yields a fixed set of cluster-based targets per image that can be reused across experiments without additional VLM calls. In contrast, VLM-as-a-judge incurs repeated inference for each new system output, so its cost scales with the number of evaluated models, prompts, or checkpoints.

\section*{Acknowledgements}
We thank the CincyNLP group for helpful discussions, and the anonymous reviewers for their valuable feedback and constructive suggestions.



\bibliography{ref}

\newpage

\appendix

\section{Prompting LLMs}
\label{app:prompt}
As we aim to evaluate the multiple perspectives of the images, one of the key challenges is obtaining comprehensive descriptions that include all perspective verbs associated with an image. To acquire this information, we employ state-of-the-art (SOTA) multimodal large language models (LLMs) to label the images. We utilize Llama-3.2-11B and Llama-3.2-90B~\citep{dubey2024llama}, both widely regarded as among the best open-source models available. Additionally, we query GPT-4o mini~\citep{hurst2024gpt} through the OpenAI API. While GPT-4o is considered superior, labeling over 100k images with it is beyond our budget constraints.

For this task, we experimented with two prompting strategies. The first is an open prompt, which asks the model to output the top 5 verbs that describe the image. The second is a closed prompt, where the 504 target verbs are provided, and the model is asked to select all verbs that describe the activity in the image. Both prompts are detailed below:

\begin{tcolorbox}[colback=gray!5!white, colframe=gray!75!black, sharp corners, breakable]
\textbf{Open Prompt:}
\begin{verbatim}
What are some verbs that describe what
is happening in this image?
Answer only with comma separated verbs
in the gerund form (they end in 'ing').
Do not include more than 5.
\end{verbatim}

\textbf{Closed Prompt:}
\begin{verbatim}
You are provided with a list of 504 
verbs: eating, walking, piloting, ...
Identify and list all verbs from this 
set that accurately describe the 
activity depicted in the image. 
Respond with the verbs only, separated 
by commas.
\end{verbatim}
\end{tcolorbox}

Since we focus on the 504 verbs in the imSitu dataset, we filter the output of the open prompt by matching it to this list. However, we observed that while the LLMs often generated reasonable replies with the open prompt, the verb distribution differed significantly. After filtering, as much as $50\%$ of the images received no valid replies within the 504 list for Llama-3.2-11B. Consequently, we use the closed prompt results as the primary gold labels for the clustering steps to identify ambiguities.

Llama-3.2-11B often produced uninformative replies, such as ``I cannot assist with this task,'' or repeated all $504$ verbs in its response. Consequently, we rely exclusively on the results from the closed prompt of Llama-3.2-90B and GPT-4o mini for the clustering steps, omitting Llama-3.2-11B due to its lower reliability.

\section{Clustering Evaluation Metrics}
\label{app:clustering_metrics}
To evaluate the quality of the clustering results, we employ three widely-used metrics: Silhouette score, Calinski-Harabasz index, and Purity score. Each metric provides a complementary view of clustering performance by quantifying different aspects of cluster quality, including cohesion, separation, and alignment with ground-truth labels. Both the Silhouette score and Calinski-Harabasz index are unsupervised metrics that measure the balance between within-cluster compactness and between-cluster separation. The key difference lies in their focus: the Silhouette score evaluates pairwise distances between points, while the Calinski-Harabasz index assesses deviations of points from their cluster centroids.

\vspace{3pt}
\noindent \textbf{Silhouette score:} The Silhouette score quantifies how well-separated and cohesive the clusters are. For a data point $i$, the Silhouette score is defined as: 
\begin{equation} 
s(i) = \frac{b(i) - a(i)}{\max(a(i), b(i))}, 
\end{equation} 
where $a(i)$ is the average distance from $i$ to all other points within the same cluster, and $b(i)$ is the smallest average distance from $i$ to points in any other cluster. In our evaluation, we compute the Silhouette score using cosine distance to better capture semantic similarities in the high-dimensional embedding space. The overall Silhouette score is the mean of $s(i)$ over all data points. Higher values, closer to 1, indicate well-separated and compact clusters.

\vspace{3pt}
\noindent \textbf{Calinski-Harabasz Index:} The Calinski-Harabasz index, also called the Variance Ratio Criterion, measures how well-defined the clusters are by comparing how far apart the clusters are to how tightly packed the points are within each cluster. It is calculated as: 
\begin{equation} 
CH = \frac{\text{trace}(B_k)}{\text{trace}(W_k)} \cdot \frac{N - k}{k - 1}, 
\end{equation} 
where $B_k$ and $W_k$ are scatter matrices that quantify the dispersion of data points. Specifically, $B_k$ (between-cluster scatter matrix) measures how spread out the cluster centroids are from the overall data centroid, capturing the separation between clusters. In contrast, $W_k$ (within-cluster scatter matrix) measures how tightly data points are grouped around their respective cluster centroids, reflecting the compactness of individual clusters. Additionally, $N$ represents the total number of data points, and $k$ denotes the number of clusters.

To make sure the distance calculations align with cosine similarity, we normalize all embeddings to have a length of 1 before clustering. A higher Calinski-Harabasz index means the clusters are well-separated from each other and the points within each cluster are tightly grouped.

\vspace{3pt}
\noindent \textbf{Purity Score:} Since there is no gold standard to evaluate our clustering results, we compute the Purity score using WordNet Synsets~\citep{miller1995wordnet}. For each cluster, we identify all synsets that contain at least one verb from the cluster and calculate the Purity score based on the maximum overlap between the cluster and any synset. To ensure meaningful evaluation, we calculate Purity only for clusters containing at least two verbs, as mapping single-verb clusters to the correct synset for comparison is ambiguous. The Purity score is defined as: \begin{equation} \text{PuS} = \frac{ \sum_{\substack{|\text{Cluster}i| > 1}} \max{j} |\text{Cluster}_i \cap \text{Synset}j| }{ \sum{\substack{|\text{Cluster}_i| > 1}} |\text{Cluster}_i| }, \end{equation} where $|\text{Cluster}_i|$ is the size of cluster $i$, and $|\text{Cluster}_i \cap \text{Synset}_j|$ represents the overlap between cluster $i$ and synset $j$. This metric approximates the semantic coherence of clusters by measuring their alignment with WordNet synsets.

\section{Supervised Classification Model}
\label{app:supervised_models}

To evaluate the performance of the classification model using our clusters, we train three different supervised classification models. We leverage three image encoders: CLIP~\citep{radford2021learning}, ResNet-50~\citep{he2016deep}, and Multimodal Llama-11B~\citep{dubey2024llama}—for verb prediction on the imSitu training set. Each backbone processes the input and generates embeddings of varying dimensions:
\begin{itemize} 
\item \textbf{ResNet-Based Model:} Uses ResNet-50, trained on ImageNet-1k (Version 1), to generate 2048-dimensional visual embeddings. 
\item \textbf{CLIP-Based Model:} Utilizes the Vision Transformer architecture, ViT-B/32 (Base, Patch Size 32), to produce 512-dimensional image embeddings. 
\item \textbf{Llama-11B-Based Model:} The model processes the input image through its vision encoder, while the text stream contains only a special \texttt{<|image|>} token. Both the image and text streams are passed through the language model using the cross-attention mechanism. The final token from the last layer serves as a 4096-dimensional embedding.
\end{itemize}
In all three models, the backbone parameters are kept frozen, and a trainable linear classification head is added on top. This classification head projects the embeddings to an output dimension of 504, representing the number of verbs in the dataset. The outputs are processed through a softmax layer and trained using cross-entropy loss, with Stochastic Gradient Descent (SGD) as the optimizer.

Early works~\citep{yatskar2016situation,pratt2020grounded} relied on CNN-based backbones for image encoding, establishing a foundation for visual activity recognition. Recent advancements have introduced CLIP-based models~\citep{li2022clip, roy2024clipsitu} and transformer-based architectures~\citep{jiang2023exploiting}, leading to significant improvements in performance. Our ResNet-based and CLIP-based models closely match the performance reported in these prior studies, as shown in Table~\ref{tab:sota_imsitu}, confirming the robustness and effectiveness of our implementations.

While achieving state-of-the-art (SOTA) performance was not the primary objective of this study, our Llama-11B-based model demonstrates strong results, achieving a Top-1 accuracy of $56\%$ on the imSitu test set. This performance approaches the current SOTA~\citep{roy2024clipsitu} for verb prediction, underscoring the potential of large-scale language models in this domain.

\begin{table}[t] 
    \centering
    \resizebox{0.98\linewidth}{!}{ 
    \begin{tabular}{lcccc}
        \toprule
        \textbf{Model} & 
        \textbf{Top1-G} & 
        \textbf{Top1-C} & 
        \textbf{Top5-G} & 
        \textbf{Top5-C} \\ 
        \midrule
        CRF~(\citealp{yatskar2016situation}) & 0.32 & - & 0.59 & - \\
        CLIP-Event~(\citealp{li2022clip}) & 0.46 & - & - & - \\
        ARF~(\citealp{jiang2023exploiting}) & 0.47 & - & - & - \\
        ClipSitu~(\citealp{roy2024clipsitu}) & 0.58 & - & 0.86 & - \\
        \midrule
        ResNet                   & 0.31 & 0.43 & 0.56 & 0.71 \\ 
        CLIP                     & 0.46 & 0.62 & 0.75 & 0.87 \\ 
        Llama                    & 0.56 & 0.72 & 0.83 & 0.92 \\ 
        \bottomrule
    \end{tabular}
    }
    \caption{Performance comparison between our supervised models and previous state-of-the-art on imSitu test set.}
    \label{tab:sota_imsitu}
\end{table}

\section{LLM-as-Judge}
\label{app:llm as judge}
Recently, LLM-as-Judge has emerged as a popular paradigm for automatic evaluation, offering scalability but raising concerns about reliability and alignment with human judgment. To assess the viability of this approach, we conducted two complementary experiments on the same 100-image subset used in our human judgment analysis, using GPT-4o and GPT-4o mini as judges. First, we evaluate model performance via an accuracy-based comparison by querying whether model predictions correctly describe the image. Second, we analyze agreement with human judgments using Kendall’s $\tau$ and Krippendorff’s $\alpha$, enabling a direct comparison of ranking consistency and label-level alignment.

\subsection{Accuracy-Based Evaluation}
\label{app:llm as judge accuracy based}
For the accuracy-based evaluation, we query the LLM judges using the following binary prompts for Top-1 and Top-5 predictions.
\begin{tcolorbox}[colback=gray!5!white, colframe=gray!75!black, sharp corners, breakable]
\textbf{Top 1:}
\begin{verbatim}
Does verb 'VERB' describe the image?
Respond with 'yes' or 'no'.
\end{verbatim}

\vspace{1em}

\textbf{Top 5:}
\begin{verbatim}
Do any verbs in 'VERB1, VERB2, ...'
describe the image?
Respond with 'yes' or 'no'.
\end{verbatim}
\end{tcolorbox}

We found that both models tended to overestimate alignment with human-labeled correctness, and their judgments varied substantially across model outputs. The discrepancy between GPT-4o and GPT-4o mini further highlights the inherent instability of LLM-as-Judge approaches in this setting.

\subsection{Agreement-Based Evaluation}
Beyond absolute accuracy, we evaluate alignment between LLM-based judges and human judgments using two complementary agreement measures. Kendall’s $\tau$ measures agreement in relative ranking between two orderings over the same set of models. Intuitively, it asks whether pairs of models are ranked in the same order by two evaluation methods. For any pair of models, the pair is considered \emph{concordant} if both rankings agree on which model performs better, and \emph{discordant} if they disagree. Kendall’s $\tau$ is then defined as
\[
\tau = \frac{N_c - N_d}{\tfrac{1}{2}n(n-1)},
\]
where $N_c$ and $N_d$ denote the numbers of concordant and discordant model pairs among $n$ models. A higher value of $\tau$ indicates stronger consistency in relative model ordering.

Krippendorff’s $\alpha$ measures \emph{label-level agreement} between LLM judgments and human annotations across individual instances. It is defined as
\[
\alpha = 1 - \frac{D_o}{D_e},
\]
where $D_o$ denotes the observed disagreement between annotators and $D_e$ denotes the expected disagreement under chance. Higher values of $\alpha$ indicate closer alignment in absolute labeling decisions.

Using the same 100-image subset, we compute both metrics for multiple runs of GPT-4o and GPT-4o mini. Specifically, rounds r1 and r2 use the same prompt as defined in Section~\ref{app:llm as judge accuracy based}, allowing us to assess variability across repeated runs. Rounds r3 and r4 introduce prompt variations to evaluate sensitivity to prompt formulation, and the corresponding prompts are shown below.

\begin{tcolorbox}[colback=gray!5!white, colframe=gray!75!black, sharp corners, breakable]
\textbf{Round 3 Prompts:}

\textbf{Top-1:}
\begin{verbatim}
Is the activity 'TOP_1' happening
in the image?
Respond with 'yes' or 'no'.
\end{verbatim}

\vspace{1em}

\textbf{Top-5:}
\begin{verbatim}
Is any activity in 'VERB1, VERB2, ...' 
happening in the image?
Respond with 'yes' or 'no'.
\end{verbatim}

\vspace{1.5em}

\textbf{Round 4 Prompts:}

\textbf{Top-1:}
\begin{verbatim}
Determine whether the verb 'TOP_1' is
visually supported by the image.
Respond with 'yes' or 'no'.
\end{verbatim}

\vspace{1em}

\textbf{Top-5:}
\begin{verbatim}
Determine whether any verb in 
'VERB1, VERB2, ...'
is visually supported by the image.
Respond with 'yes' or 'no'.
\end{verbatim}
\end{tcolorbox}
This design enables a direct analysis of stability in LLM-as-Judge evaluations and facilitates comparison with our cluster-based evaluation in terms of both ranking consistency and label-level reliability.

\subsection{Discussion}
Our results show that off-the-shelf LLM-as-Judge both overestimates model accuracy relative to human judgments and exhibits substantial instability under prompt perturbations, with agreement varying across repeated runs and prompt formulations. While more advanced prompting techniques, such as Chain-of-Thought or in-context examples, may improve reliability, our clustering approach provides a more stable and interpretable alternative by producing explicit verb sense groupings that support both dataset-level and instance-level analysis. In contrast, LLM-as-Judge methods yield binary or probabilistic judgments without an explicit semantic structure, a limitation that has been increasingly noted in recent studies~\cite{chen2024mllm, thakur2024judging}.

\section{Error Analysis}
\label{app:error analysis}
\begin{figure*}[t]
    \centering
    \includegraphics[width=0.95\linewidth]{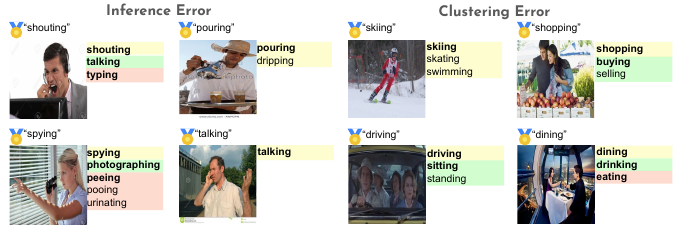}
    \caption{Error analysis of our clustering pipeline. Inference predictions are shown in \textbf{bold}, and different clusters for the same image are distinguished by background colors.}
    \label{fig:real error analysis}
\end{figure*}
While our clusters have proven effective in capturing verb semantics, marginal errors still occur throughout the pipeline. We conduct an error analysis by categorizing these errors into two main types: \textit{inference error} and \textit{clustering error}, with illustrative examples shown in Figure~\ref{fig:real error analysis}. Each example includes the gold label, inference predictions (in \textbf{bold}), and other verbs within the same cluster. Different clusters corresponding to the same image are distinguished by background color.

Inference error arises during the generation of \texttt{<image,verb>} pairs and typically occurs when the predicted activity does not accurately or completely reflect the image content. As shown in the first column of Figure~\ref{fig:real error analysis}, the verb \textit{typing} is an incorrect prediction for the image labeled \textit{shouting}, and while \textit{photographing} may be loosely related to the image labeled \textit{spying}, the inclusion of \textit{peeing} is clearly irrelevant.
Another form of inference error occurs when plausible activities are omitted. As shown in the second column, \textit{serving} would be an equally valid label for the \textit{pouring} image, and the man in the \textit{talking} image is clearly also \textit{phoning}.

Clustering error is also two-fold. The first type occurs when semantically unrelated or loosely related activities are grouped together. As shown in the third column of Figure~\ref{fig:real error analysis}, while \textit{skiing} and \textit{skating} are clearly related, \textit{swimming}, though also a sport, is less relevant. Similarly, \textit{sitting} and \textit{standing} often co-occur but are semantically distinct: for instance, \textit{sitting} can describe the \textit{driving} image, while \textit{standing} cannot.
The second type of clustering error involves cases where multiple related clusters could be merged. Although this type is less critical for evaluation, we show examples in the last column. In the \textit{shopping} image, \textit{shopping} and \{\textit{buying}, \textit{selling}\} are closely related and frequently co-occur, suggesting they could belong to the same cluster. Similarly, \textit{dining} and \textit{eating} could be grouped together to better describe the \textit{dining} image.

\section{Analysis on CLIPScore}
\label{app:clipscore}
This section presents further details regarding the CLIPScore analysis.

\vspace{3pt}
\noindent
\textbf{With context:} To perform the evaluation, we generate both correct and modified captions for each image. In the imSitu dataset, each verb is associated with an abstract template and a predefined set of semantic roles. For example, the verb \textit{soaring} is linked to the abstract template \textit{an AGENT soars in a PLACE}, where \textit{AGENT} and \textit{PLACE} denote semantic roles. These roles are annotated for each image labeled with the corresponding verb. To obtain the correct caption, we prompt GPT-4o mini to generate a natural language sentence. An example prompt is shown below:
\begin{tcolorbox}[colback=gray!5!white, colframe=gray!75!black, sharp corners, breakable]
\begin{verbatim}
Given the abstract structure 'an AGENT 
soars in a PLACE' and the following
components: 
'place': 'sky', 
'agent': 'bird', 
generate a natural sentence. 
Reply with only the sentence.
\end{verbatim}
\end{tcolorbox}
After obtaining the correct caption, we generate modified captions using two approaches. The first involves replacing the verb in the correct caption with a randomly selected verb from the imSitu verb list. The second approach prompts GPT-4o to substitute the verb with an implausible alternative. The prompt used for this transformation is as follows:
\begin{tcolorbox}[colback=gray!5!white, colframe=gray!75!black, sharp corners, breakable]
\begin{verbatim}
You are given the following sentence: 
'{correct caption}'
Can you change the verb in the sentence 
to an unreasonable one without 
changing other parts of the sentence?
Reply with only the new sentence.
\end{verbatim}
\end{tcolorbox}
Once both the correct and modified captions are generated, we compute the cosine similarities between their CLIP embeddings. The results are presented in Table~\ref{tab:clipscore caption}.

\begin{table}[ht]
  \centering
  \begin{tabular}{lcc}
    \toprule
          & \textbf{Top 1} & \textbf{Top 5} \\
    \midrule
    Train & 0.25 & 0.48 \\
    Test  & 0.26 & 0.49 \\
    Dev   & 0.26 & 0.49 \\
    \bottomrule
  \end{tabular}

  \caption{Accuracy on the imSitu dataset using CLIPScore for verb ranking.}
  \label{table: clipscore}
\end{table}

\vspace{3pt}
\noindent
\textbf{Without context:} To further study the ability of CLIPScore for evaluating image-activity alignment in the absence of context, we compute CLIP similarity scores between each image and all 504 verbs in the imSitu dataset. The verbs are then ranked by their CLIPScore, and we assess Top-1 and Top-5 prediction accuracy against the gold-standard labels. The results, presented in Table~\ref{table: clipscore}, show that CLIPScore achieves a Top-1 accuracy of approximately 25\% (i.e., 25\% of the images receive their highest similarity score with the gold-standard verb), while the Top-5 accuracy remains below 50\%. These results are consistent across the train, dev, and test splits.




\section{Clustering Algorithm}
\label{app:clustering}
\begin{algorithm*}[ht]
\caption{Two-Step Clustering Framework with Silhouette Optimization}
\label{alg:clustering}
\begin{algorithmic}[1]
\State \textbf{Input:} A dataset of \texttt{<image,verb>} pairs $\mathcal{D}$, a list of 504 target verbs $\mathcal{V}$, a clustering algorithm $\mathcal{A}$, a candidate list of cluster numbers $\mathcal{K}$ for Step 1, and a candidate list of ratios $\mathcal{R}$ for Step 2.
\State \textbf{Output:} A set of final clusters $\mathcal{C}_{\text{final}}$.

\State \textbf{Step 1: Same-Verb Clustering}
\For{each verb $v \in \mathcal{V}$}
    \State Collect all images $\mathcal{I}_v$ associated with $v$ from $\mathcal{D}$.
    \State Generate embeddings $\mathcal{E}_v$ for $\mathcal{I}_v$ using a multimodal language model.
    \State Initialize $\text{best\_score} \gets -\infty$ and $\text{best\_clusters} \gets \emptyset$.
    \For{each $k \in \mathcal{K}$}
        \State Apply $\mathcal{A}$ to $\mathcal{E}_v$ with $k$ clusters to obtain clustering result $\mathcal{C}_v^{(k)}$.
        \State Compute the Silhouette score $\text{silhouette}(\mathcal{C}_v^{(k)})$.
        \If{$\text{silhouette}(\mathcal{C}_v^{(k)}) > \text{best\_score}$}
            \State Update $\text{best\_score} \gets \text{silhouette}(\mathcal{C}_v^{(k)})$.
            \State Update $\text{best\_clusters} \gets \mathcal{C}_v^{(k)}$.
        \EndIf
    \EndFor
    \State Save $\text{best\_clusters}$ as the optimal clustering for $v$.
\EndFor
\State Store all optimal clusters from Step 1 as intermediate clusters $\mathcal{C}_{\text{step1}}$.

\State \textbf{Step 2: Cross-Verb Clustering}
\State Compute the total number of verbs as $|\mathcal{V}| = 504$.
\State Initialize $\text{best\_score} \gets -\infty$ and $\text{final\_clusters} \gets \emptyset$.
\For{each ratio $r \in \mathcal{R}$}
    \State Compute candidate cluster number $k_r = \text{int} (504 \times r)$.
    \State Apply $\mathcal{A}$ to the average embeddings of $\mathcal{C}_{\text{step1}}$ with $k_r$ clusters to obtain clustering result $\mathcal{C}_{\text{final}}^{(k_r)}$.
    \State Compute the Silhouette score $\text{silhouette}(\mathcal{C}_{\text{final}}^{(k_r)})$.
    \If{$\text{silhouette}(\mathcal{C}_{\text{final}}^{(k_r)}) > \text{best\_score}$}
        \State Update $\text{best\_score} \gets \text{silhouette}(\mathcal{C}_{\text{final}}^{(k_r)})$.
        \State Update $\text{final\_clusters} \gets \mathcal{C}_{\text{final}}^{(k_r)}$.
    \EndIf
\EndFor

\State \textbf{Return:} $\text{final\_clusters}$.
\end{algorithmic}
\label{algo: clustering}
\end{algorithm*}

\newpage

\end{document}